\documentclass[letterpaper]{article} 

\usepackage{aaai2026}  

\usepackage{times}  
\usepackage{helvet}  
\usepackage{courier}  
\usepackage[hyphens]{url}  
\usepackage{graphicx} 
\usepackage{natbib}  
\usepackage{caption} 

\usepackage[utf8]{inputenc} 
\usepackage[T1]{fontenc}    
\usepackage{url}            
\usepackage{amsmath}            
\usepackage{booktabs}       
\usepackage{amsfonts}       
\usepackage{nicefrac}       
\usepackage{microtype}      
\usepackage{xcolor}         
\usepackage{graphicx}
\usepackage{multirow}
\usepackage{float}
\usepackage{caption} 
\usepackage{array}
\usepackage{makecell}
\usepackage{algorithm}
\usepackage{algorithmic}
\usepackage{tcolorbox}
\usepackage{inconsolata}
\usepackage{xcolor}
\usepackage{bibentry}
\usepackage{newfloat}
\usepackage{listings}
\tcbuselibrary{breakable}
\DeclareCaptionStyle{ruled}{labelfont=normalfont,labelsep=colon,strut=off} 
\floatstyle{ruled}
\newfloat{listing}{tb}{lst}{}
\floatname{listing}{Listing}
\definecolor{promptbg}{rgb}{0.95,0.95,0.98}
\definecolor{promptframe}{rgb}{0.7,0.7,0.8}
\title{GEM: Graph-Enhanced Mixture-of-Experts with ReAct Agents for Dialogue State Tracking}

\author{
    Ziqi Zhu\thanks{These authors contributed equally.},
    Adithya Suresh\footnotemark[1],
    Tomal Deb\thanks{These authors contributed equally.},
    Iman Abbasnejad\footnotemark[2]
}
\affiliations{
    Generative AI Innovation Center, Amazon Web Services\\
    \{ziqizhu, adxthya, tomalde, imanaba\}@amazon.com
}

\begin{document}

\maketitle

\begin{abstract}

  Dialogue State Tracking (DST) requires precise extraction of structured information from multi-domain conversations, a task where Large Language Models (LLMs) struggle despite their impressive general capabilities. We present \textbf{GEM} (Graph-Enhanced Mixture-of-Experts), a novel framework that combines language models and graph-structured dialogue understanding with ReAct agent-based reasoning for superior DST performance. Our approach dynamically routes between specialized experts: a Graph Neural Network that captures dialogue structure and turn-level dependencies, and a finetuned T5-Small encoder-decoder for sequence modeling, coordinated by an intelligent router. For complex value generation tasks, we integrate ReAct agents that perform structured reasoning over dialogue context. On MultiWOZ 2.2, GEM achieves \textbf{65.19\% Joint Goal Accuracy}, substantially outperforming end-to-end LLM approaches (best: 38.43\%) and surpassing state-of-the-art (SOTA) methods including TOATOD (63.79\%), D3ST (58.70\%), and Diable (56.48\%). Our graph-enhanced mixture-of-experts architecture with ReAct integration demonstrates that combining structured dialogue representation with dynamic expert routing and agent-based reasoning provides a powerful paradigm for dialogue state tracking, achieving superior accuracy while maintaining computational efficiency through selective expert activation.

\end{abstract}

\section{Introduction}

Dialogue State Tracking (DST) constitutes a fundamental component of task-oriented dialogue systems, responsible for accurately monitoring user intentions and extracting structured information throughout conversational interactions. As conversations progress across multiple turns and domains, DST systems must maintain precise representations of user goals, preferences, and constraints to enable appropriate system responses. Despite significant advances in neural architectures~\cite{wu2019transferable, bebensee-lee-2023-span, Lesci_2023}, achieving robust performance in complex multi-domain scenarios remains challenging due to the inherent difficulties in tracking evolving contextual information and resolving ambiguities across conversation turns.
\begin{figure}[t]
    \centering
    \includegraphics[width=0.45\textwidth]{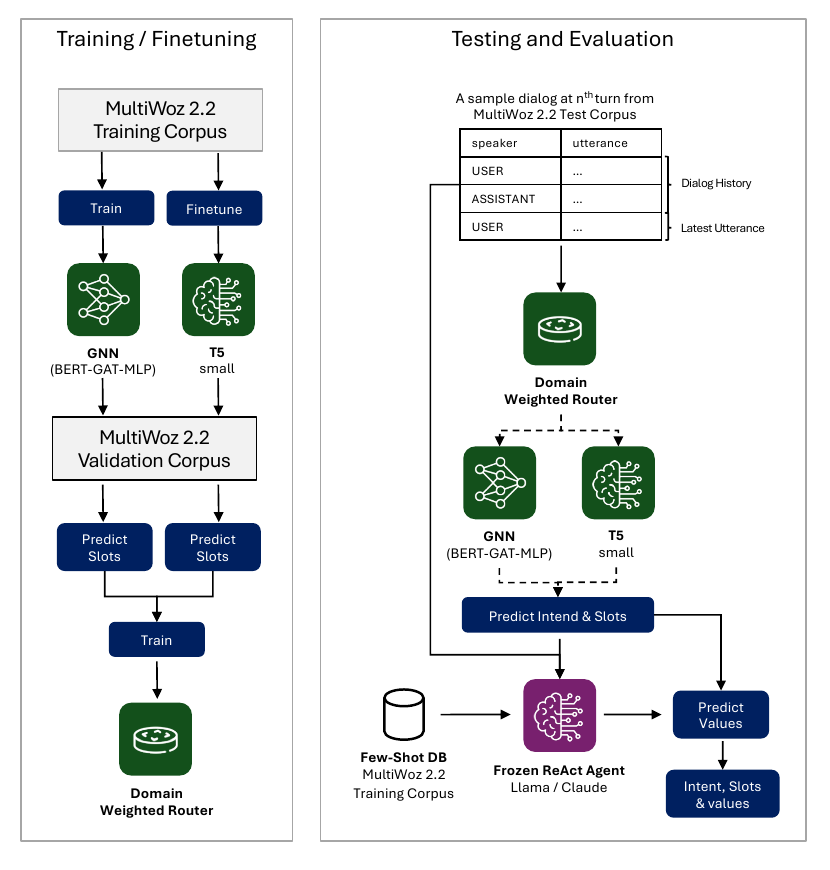}
    \caption{Our proposed architecture for dialogue state tracking. The MoE dynamically routes inputs between T5 encoder-decoder~\cite{raffel2020exploring} and GNN model via a learned gating network. This hybrid approach effectively combines the strengths of transformer-based language understanding with graph-based relational reasoning for improved intent detection, slot prediction, and value extraction in multi-turn conversations.}
    \label{fig:workflow}
    \vspace{-1em}
\end{figure}

The development of large language models (LLMs) has significantly transformed natural language processing capabilities, demonstrating unprecedented performance across diverse tasks~\cite{mann2020language, bang2023multitask}. Particularly noteworthy is their emergent reasoning ability, which enables complex problem-solving through stepwise decomposition of tasks~\cite{wei2022chain, huang2022towards}. The plan-and-solve paradigm~\cite{chen2025improving} has further enhanced this capacity by prompting LLMs to generate explicit reasoning steps, allowing them to address intricate challenges through systematic deconstruction into manageable sub-tasks~\cite{khot2022decomposed}. However, as these models grow increasingly large and complex, they present substantial computational challenges while still exhibiting fundamental limitations in knowledge representation and reasoning reliability. Despite their significant advancements, LLMs remain constrained by knowledge limitations and susceptibility to hallucinations during reasoning processes, leading to consequential errors in critical applications~\cite{hong2023faithful, wang2023knowledge}.

Knowledge graphs (KGs) offer a promising direction for addressing these limitations by providing structured factual representations that can enhance dialogue understanding~\cite{pan2024unifying}. Existing integration approaches are typically divided between semantic parsing methods that convert questions into logical queries~\cite{lan2020query, ye2021rng} and retrieval-augmented methods that enhance context with relevant information~\cite{jiang2022unikgqa}. More sophisticated knowledge fusion techniques like retrieval-augmented generation (RAG)\cite{lewis2020retrieval} and knowledge graph prompting\cite{zhang2024knowgpt} have emerged to address these limitations. However, these approaches face persistent challenges: semantic parsing often produces non-executable queries due to syntactic constraints, while retrieval-based methods typically fail to fully utilize the rich structural information inherent in knowledge graphs~\cite{mavromatis2024gnn}. Furthermore, agent-based methods necessitate multiple interaction rounds, increasing computational overhead~\cite{dehghan2024ewek}.

Computational efficiency presents another critical consideration for advanced dialogue systems. The Mixture-of-Experts (MoE) approach~\cite{xue2024openmoe} offers a promising solution by activating only a subset of parameters or models at any given time, allowing the total parameter count to grow without proportional increases in computational requirements. LLMs scaled using this approach have demonstrated impressive performance across downstream tasks~\cite{jiang2024mixtral}, yet their application to structured dialogue state tracking remains underexplored.

This paper introduces a novel framework for DST that addresses these challenges through a Graph Enhanced Mixture-of-Experts (GEM) architecture. Our approach dynamically constructs and updates knowledge graphs through graph neural networks (GNNs), harnessing their representation learning capabilities to process complex structural information while overcoming their inherent limitations in natural language understanding. Different dialogue domains naturally exhibit distinct processing requirements — some benefit from structural reasoning while others require sequential linguistic understanding — motivating our dynamic expert routing approach. The framework employs a routing mechanism that dynamically allocates computational resources between a finetuned T5 encoder-decoder model~\cite{raffel2020exploring} and our proposed GNN-based method, activating only the most appropriate expert for each specific dialogue segment. Unlike conventional methods, our graph-based representation system captures the intricate, evolving relationships between dialogue states, user intents, and system responses throughout conversation turns. We further enhance the architecture through an agentic-based approach for value generation that leverages structured reasoning via the ReAct framework, significantly improving both the accuracy and interpretability of slot-value extraction in challenging multi-domain conversational scenarios. Figure~\ref{fig:workflow} illustrates our proposed method.

In this paper we make the following contributions:
\begin{enumerate}
\vspace{-5pt}
\item A fine-tuned graph attention network architecture for accurate intent detection and slot prediction that effectively models complex interactional dynamics in multi-turn conversations,
\vspace{-5pt}
\item A routing mechanism that dynamically selects between finetuned T5 encoder-decoder and GNN expert models based on query characteristics, optimizing the balance between computational efficiency and prediction accuracy,
\vspace{-5pt}
\item Extending the architecture by integration of a pre-trained decoder-LLM for robust, slot-conditioned value extraction,
\vspace{-5pt}
\item Extending the solution through an agentic-based approach that leverages conversation history, content context, and identified slots to perform reasoning-based value generation,
\vspace{-5pt}
\item Obtaining the SOTA results on the challenging MultiWOZ dataset, demonstrating the framework's effectiveness in complex dialogue systems.
\end{enumerate}

\section{Related work}
DST is a critical component of task-oriented dialogue systems, requiring precise tracking of user intents and slot values throughout conversations. Traditional approaches have evolved from rule-based systems to neural network architectures, with sequence-based models like TRADE \cite{wu2019transferable} and TripPy \cite{heck2020trippy} establishing strong baselines. More recent innovations include DS-DST~\cite{zhang2019find}, Seq2Seq-DU~\cite{feng2021sequencetosequenceapproachdialoguestate}, LUNA~\cite{wang-etal-2022-luna}, SPACE-3~\cite{he2022space3unifieddialogmodel}, SPLAT~\cite{bebensee-lee-2023-span}, Diable~\cite{Lesci_2023}, D3ST \cite{zhao2022description} and TOATOD \cite{bang2023task}. While these models have advanced performance boundaries, they continue to face limitations in modeling complex interactional dynamics across multi-turn dialogues, particularly in multi-domain scenarios where context flows between distinct conversational topics.

LLMs have transformed the DST landscape, demonstrating impressive general language understanding capabilities \cite{mann2020language, bang2023multitask}. However, they faces challenges in extracting structured information and maintaining consistency, leading to issues like hallucination \cite{hudecek-dusek-2023-large, heck2023chatgpt, hong2023faithful}. Recent prompt-based approaches like IC-DST \cite{hu2022context}, SERI-DST \cite{lee2024inference}, InstructTODS \cite{chung2023instructtods}, and FNCTOD \cite{li2024large} leverage in-context learning but still struggle to translate broad language understanding into precise, structured dialogue states—particularly in multi-domain scenarios.

Knowledge representation frameworks offer promising solutions \cite{pan2024unifying}, through either semantic parsing \cite{lan2020query, ye2021rng} or retrieval-augmented approaches \cite{lewis2020retrieval, jiang2022unikgqa}, though each approach has limitations. While more sophisticated knowledge fusion techniques like RAG \cite{lewis2020retrieval} and KG prompting \cite{zhang2024knowgpt} have demonstrated improved performance, they still face challenges with retrieval accuracy and computational efficiency \cite{mavromatis2024gnn, dehghan2024ewek}. GNNs \cite{gori2005new, velivckovic2017graph, mavromatis2025gnn, luo2025gfm} show promise for modeling complex relational structures, yet their application to dialogue understanding remains underdeveloped. Existing graph-based DST methods \cite{chen2020schema} often rely on static, predefined ontologies that cannot dynamic dialogue evolution.

Computational efficiency presents another significant challenge, particularly as models grow more complex to handle sophisticated dialogue understanding tasks. MoE architectures \cite{shazeer2017outrageously, xue2024openmoe} have demonstrated parameter efficiency in large-scale language models \cite{jiang2024mixtral}, though their application to DST remains unexplored. Similarly, agent-based approaches like ReAct \cite{yao2023react} have shown promise for enhancing reasoning capabilities through explicit stepwise decomposition \cite{wei2022chain, khot2022decomposed}, but require further investigation for structured DST.

These limitations highlight the need for a hybrid architecture that combines graph-based dialogue representation for structural understanding, efficient computational allocation through expert routing, and specialized value generation for accurate knowledge extraction. Our work addresses these challenges through the integration of GNN, MoE, and specialized state value generators within a unified dialogue state tracking framework.

\section{Method}
\label{sec:method}
We present a hybrid architecture that combines Graph Neural Networks (GNNs) and sequence model for structural dialogue understanding, Mixture of Experts (MoE) for efficient computation allocation, and state value generators for accurate knowledge extraction.
\subsection{Intents and Slots Detection}
We use two approaches for intent and slot detection, (i) a GNN that captures structural relationships across dialogue turns, and (ii) a T5 encoder-decoder model that leverages sequential context for comprehensive language understanding.

\subsubsection{Graph Neural Network}
\label{sec:gnn}
Following the success of GNNs\cite{gori2005new} in processing graph-structured data, we construct our dialogue understanding framework using the Graph Attention Network (GAT)\cite{velivckovic2017graph, brody2021attentive} to effectively model the complex interactional dynamics presentation in multi-turn conversations. Our approach leverages GAT's attention mechanism specifically for joint slot filling and intent detection tasks in dialogue understanding. In this work, we represent dialogue as graph structure $G = (V, E)$, where nodes $v_i \in V$ contains the contextual embedding of utterances, with special tokens [USER] and [ASSISTANT] to distinguish speaker types. The edges $E$ create directed temporal connections between consecutive utterances, enabling information flow for both intent classification and slot labeling tasks.

For both slot filling and intent detection, we compute node representations through $L$ layers of graph attention. At layer $l$, the representation is updated as:
\begin{align}
h_i^{(l+1)} &= \sigma\left(\sum_{j \in \mathcal{N}_{i}} \alpha_{ij}^{l} W^{(l)} h_j^{(l)}\right),
\end{align}

where, $\mathcal{N}_{i}$ is the neighborhood of node $i$, $W^{(l)} \in \mathbb{R}^{F' \times F}$ is a learnable weight matrix, $\alpha_{ij}^{(l)}$ are attention coefficients computed as:
\begin{equation}
\alpha_{ij}^{(l)} = \frac{\exp\left(\text{LeakyReLU}\left(\vec{a}^T[W^{(l)}h_i^{(l)} | W^{(l)}h_j^{(l)}]\right)\right)}{\sum_{k \in \mathcal{N}_{i}} \exp\left(\text{LeakyReLU}\left(\vec{a}^T[W^{(l)}h_i^{(l)} | W^{(l)}h_k^{(l)}]\right)\right)},
\end{equation}
where $|$ represents concatenation and $\vec{a}$ is a learnable attention vector.

For both slot filling and intent detection tasks, we perform utterance-level multi-label classification using the final node representations. We employ $K$ independent attention heads and combine their outputs:
\begin{equation}
h_i^{(L)} = \sigma\left(\frac{1}{K}\sum_{k=1}^K \sum_{j \in \mathcal{N}i} \alpha_{ij}^{k,(L-1)} W_k^{(L-1)}h_j^{(L-1)}\right).
\end{equation}

The multi-head attention mechanism helps the model attend to different aspects of the dialogue context, which is particularly beneficial for capturing the relationships between slots and intents across multiple turns. This approach allows our model to jointly learn slot and intent representations while accounting for the dialogue history and conversational dynamics.

\subsubsection{T5 Encoder-Decoder Model} 
\label{sec:t5} 
While graph-based approaches excel at modeling structural relationships, transformer-based sequence models have demonstrated exceptional capabilities in natural language understanding tasks~\cite{raffel2020exploring}. The T5 architecture models intent detection and slot filling as a sequence-to-sequence generation task, where the input is the dialogue context and the output is a structured representation of intents and slots. We fine-tune the T5 model using a standard sequence-to-sequence objective that maximizes the probability of generating the correct structured output given the dialogue context.

\subsection{State Value Generator}
After slot and intent detection, we extract slot values using two approaches that share a common dynamic few-shot example selection mechanism.

\subsubsection{Dynamic Few-Shot Example Selection} 
\label{sec:fewshot}
To enable effective information retrieval for few-shot learning, we create dense vector representations of dialogue contexts. Each training example is encoded using Cohere's embedding model~\cite{cohere_embed}, represented as $\mathbf{z}_{i} = \text{Embed}(f_{\text{combine}}(D^{\text{sys}}_{i-1}, D^{\text{user}}i, S_i))$, where $f_{\text{combine}}$ constructs a structured text representation combining the previous system response, current user utterance, and identified slots. For a given dialogue turn $t$, we retrieve the most relevant examples by computing semantic similarity:

\begin{equation}
E_t = \text{TopK}({\mathbf{e} \in \mathcal{E} : \cos(\mathbf{z}_e, \mathbf{z}_t) > \tau_{\text{sim}}}),
\end{equation}

where $\mathbf{z}_t$ represents the embedding of the current turn and $\tau_{\text{sim}}$ is a similarity threshold. These embeddings are indexed in ChromaDB for efficient similarity-based retrieval. Each retrieved example $e \in E_t$ includes its ground truth slot $s_{i}$, value $v_{j}$ pairs $V_e = \{(s_j, v_j)\}_{j=1}^{m_e}$, providing demonstration instances for the subsequent value generation methods.

\subsubsection{LLM-Based Generation}
We utilize pre-trained decoder-only LLMs for slot value extraction through few-shot learning. For turn $t$, we construct input using dynamically retrieved examples $E_t$ along with dialogue history and identified slots to generate structured slot-value pairs:
\begin{align}
\mathbf{x}_t &= f_{\text{concat}}(\boldsymbol{\delta}_{\text{prefix}}, E_t, \boldsymbol{\delta}_{\text{history}}, D_{0:t-1}, \nonumber \\
&\quad \boldsymbol{\delta}_{\text{current}}, D_t, \boldsymbol{\delta}_{\text{slots}}, S_t),
\end{align}
\begin{equation} 
V_t = \text{LLM}_{\theta}(\mathbf{x}_t) = \{(s_i, v_i)\}_{i=1}^k,
\end{equation}
where $S_t$ are identified slots from the GNN module, $\boldsymbol{\delta}_{\cdot}$ represent delimiter tokens, and the model generates structured JSON responses with slot-value pairs.

\subsubsection{ReAct Agent-Based Generation}
\label{sec:react}
The ReAct agent utilizes structured reasoning to extract slot values through explicit analysis and reasoning steps in a single turn. Using the same retrieved examples $E_t$ as the LLM approach, the agent performs compositional reasoning functions:
\begin{align}
V_t &= \mathcal{A}_{\boldsymbol{\phi}}(E_t, D_{0:t}, S_t) \nonumber \\
&= (f_{\text{analysis}} \circ f_{\text{reasoning}} \circ f_{\text{json}})(E_t, D_{0:t}, S_t),
\end{align}
where $f_{\text{analysis}}$ analyzes dialogue context with retrieved examples, $f_{\text{reasoning}}$ provides explicit reasoning traces for identified values, and $f_{\text{json}}$ generates structured JSON outputs. This approach enhances interpretability while maintaining the same extraction protocols as the LLM-based method.

\subsection{Mixture of Experts}
\label{sec:mixture_of_experts}
To speed up learning and improve generalization across different scenarios~\cite{jacobs1991adaptive} proposed using several different expert networks instead of a single model. Mixture of expert models have been extensively studied since then, consistently demonstrating both performance improvements and cost reductions~\cite{yao2009hierarchical, liu2024survey, vats2024evolution}. In this work we consider using gating strategy for model selection. When designing an effective gating function, two key criteria must be prioritized: (1) accurate discernment of both input and expert characteristics to enable assignment of similar data to the same expert, and (2) even distribution of input data among experts to prevent model collapse and ensure efficient utilization of all experts. Our routing architecture leverages domain-weighted voting strategy to direct dialogue turns to the most appropriate expert model. The system employs a BERT-based multi-label domain classifier trained on conversational history to identify active domains for each dialogue turn. The voting scheme for our proposed method is presented in Algorithm~\ref{algo:domain-routing}, where in this algorithm $H$ represents the dialogue history, $U$ is the current utterance, Acc$_{GNN}$ and Acc$_{T5}$ represent slot accuracies for respective models.

\begin{algorithm}[t]
  \caption{Routing Mechanism}
  \label{algo:domain-routing}
  \begin{algorithmic}[1]
    \REQUIRE $H$, $U$
    \ENSURE $M \in ({\text{GNN}, \text{T5}})$
    \STATE $D \leftarrow \text{DomainClassifier}(H, U)$
    \STATE votes$_{GNN} \leftarrow 0$, votes$_{T5} \leftarrow 0$
    \FOR{each domain $d \in D$}
      \IF{Acc$_{GNN}$ $>$ Acc$_{T5}$}
        \STATE votes$_{GNN}$ $\leftarrow$ votes$_{GNN} $ $+ 1$
      \ELSE
        \STATE votes$_{T5}$ $\leftarrow$ votes$_{T5}$ $+ 1$
      \ENDIF
    \ENDFOR
    \IF{votes$_{GNN}$ $\geq$ votes$_{T5}$}
      \RETURN GNN
    \ELSE
      \RETURN T5
    \ENDIF
  \end{algorithmic}
\end{algorithm}

\section{Experiments}
We evaluate our method for dialogue state tracking task on the MultiWOZ 2.2 dataset. Detailed ablation studies examining the impact of different GNN configurations, embedding models, context window sizes and loss weighting are provided in their respective sections. Latency analysis across different model configurations is presented separately as well.
\\
\textbf{Dataset:} We conduct experiments on MultiWOZ 2.2~\cite{budzianowski2018multiwoz}, a multi-domain task-oriented dialogue dataset containing 10,438 human-human conversations across 7 domains (restaurant, hotel, attraction, taxi, train, hospital, police). This dataset provides rich annotations for intent detection and slot filling tasks, with 13 intent types and 25 slot types across approximately 13.7 average turns per dialogue. We specifically choose MultiWOZ 2.2 as it offers consistent annotation quality for intent and slot labels throughout the corpus, while later versions (2.3-2.4) primarily address co-reference annotations and test set corrections that are orthogonal to our focus. The dataset's multi-turn nature (with dialogues extending up to 30+ turns) makes it particularly suitable for evaluating contextual understanding capabilities. Additionally, MultiWOZ 2.2 remains a widely-used version in the dialogue understanding literature, enabling direct comparison with existing baselines and SOTA methods.
\\
\textbf{Evaluation Metrics:} We report Joint Goal Accuracy (JGA) and Joint Turn Accuracy (JTA) for the slot-value generation. For intent and slot classification we report turn-level accuracy.

\subsection{Experimental setup}
Our experimental framework utilizes the multi-expert architecture described in Section~\ref{sec:method}. For the GNN, we utilize BERT-base~\cite{devlin2019bert} to encode dialog histories with speaker tokens, which then feeds into a three-layer GATv2 network with three parallel MLP decoders to simultaneously predict intent, domain, and slot values, jointly optimized using binary cross-entropy loss. We use QLoRA-tuned T5-Small~\cite{raffel2020exploring} to transform concatenated dialog histories into structured domain-intent-slot combinations through sequence-to-sequence generation. For slot value extraction, we applied two approaches: (1) a retrieval-augmented generation (RAG) system and (2) a ReAct agent-based for slot-value generation. For both we use Llama 3.1 8B Instruct~\cite{grattafiori2024llama} and Claude 3.7 Sonnet~\cite{sonnet_aws} as the LLM model. All models were trained on NVIDIA A10G GPU with 24GB memory, 8 vCPUs, 32GB RAM.
\begin{table}[t]
\centering
\small
\renewcommand{\arraystretch}{1.45}
\setlength{\tabcolsep}{3pt}
\begin{tabular}{l|l|c|c}
\hline
\textbf{Model} & \textbf{Configuration} & \textbf{Intent Acc} & \textbf{Slot Acc} \\
\hline
T5 & Small & 86.69 & 73.76 \\
\hline
\multirow{4}{*}{GNN} & BERT$_{base}$ + GNN$_{base}$& $86.84$ & $71.07$ \\
& BGE $_{small}$ + GNN$_{base}$ & $87.00$ & $69.71$ \\
& BERT $_{base}$ + GNN$_{large}$ & $86.63$ & $71.55$ \\
& (BERT $_{base}$ + GNN$_{large}$)$_{opt}$ & $86.53$ & \textbf{74.54} \\
\hline
\end{tabular}
\caption{Test set performance for intent classification and slot identification across different model configurations. The GNN rows show variations in embedding models (BERT$_{base}$, BGE$_{small}$), architecture (GNN$_{base}$, GNN$_{large}$), and other optimization approaches ($opt$) including residual connections, BCE loss weighting and dialogue context window. Results demonstrate the impact of different design choices on model performance.}
\label{tab:intent_slot_perf}
\vspace{-1em}
\end{table}
\subsection{Results}
We evaluate our proposed framework on the MultiWOZ 2.2 benchmark dataset, analyzing performance across multiple dimensions: intent and slot detection accuracy, value generation quality with different LLM configurations, domain-specific routing behavior, and overall dialogue state tracking performance compared to SOTA baselines. All experiments were conducted on the standard test split to ensure fair comparison with existing methods.
\\
\textbf{Intent and Slot Detection:}
Table~\ref{tab:intent_slot_perf} provides a detailed analysis of our model's performance on intent classification and slot identification tasks across different architectural configurations. Our GNN architecture demonstrates strong performance on both intent classification ($86.53\%$) and slot identification ($74.54\%$) when using the optimized configurations. The finetuned T5-Small encoder-decoder model shows comparable intent and slot classification performance ($86.69\%$ and $73.76\%$ respectively). Ablation studies are provided in Appendix.
\\
\textbf{Dialog State Tracking using LLM:}
We evaluate three LLM configurations: zero-shot (no examples), few-shot (with in-context examples), and ReAct (reasoning and acting framework) across two LLM families. The results for this experiment are illustrated in Table~\ref{tab:llm_performance}. The ReAct based method dramatically improves Llama 3.1 8B performance, achieving $26.74\%$ JGA improvement over zero-shot prompting and representing the best end-to-end LLM results on the test set.
\begin{table}[t]
\centering
\small
\renewcommand\arraystretch{1.1}
\begin{tabular}{>{\centering\arraybackslash}p{1.7cm}>{\centering\arraybackslash}p{0.8cm}|>{\centering\arraybackslash}p{.6cm}>{\centering\arraybackslash}p{.6cm}}
\hline
\textbf{Model} & \textbf{Type} & \textbf{JGA} & \textbf{JTA} \\
\hline
Llama 3.1  8B & Zero & 10.33 & 80.65 \\
Llama 3.1  8B & Few & 32.62 & 89.77 \\
Llama 3.1  8B & ReAct & \textbf{37.07} & 93.36 \\
Claude 3.7 S. & Zero & 32.50 & 90.82 \\
Claude 3.7  S. & Few & \textbf{38.43} & 91.48 \\
Claude 3.7  S. & ReAct & 37.09 & 91.48 \\
\hline
\end{tabular}
\caption{End-to-end LLM performance on MultiWOZ 2.2}
\label{tab:llm_performance}
\end{table}

\begin{table}[t]
\centering
\small
\renewcommand\arraystretch{1.1}
\begin{tabular}{>{\centering\arraybackslash}p{1.4cm}|>{\centering\arraybackslash}p{1.0cm}>{\centering\arraybackslash}p{0.85cm}>{\centering\arraybackslash}p{1.1cm}>{\centering\arraybackslash}p{0.7cm}>{\centering\arraybackslash}p{0.5cm}}
\hline
\textbf{Metric} & \textbf{Attraction} & \textbf{Hotel} & \textbf{Restaurant} & \textbf{Taxi} & \textbf{Train} \\
\hline
GNN \newline Routing \% & \textbf{91.31} & 10.01 & 14.18 & 12.73 & \textbf{93.48} \\
T5 \newline Routing \% & 8.69 & \textbf{89.99} & \textbf{85.82 }& \textbf{87.27} & 6.52 \\
\hline
Router \newline Accuracy & \textbf{70.21} & \textbf{56.05} & \textbf{71.64} & \textbf{61.62} & \textbf{78.47} \\
\hline
\end{tabular}
\caption{MoE Performance by domain showing routing distribution and accuracy metrics for each expert model.}
\label{tab:moe_domain}
\vspace{-1em}
\end{table}

\begin{table*}[t]
\centering
\begin{minipage}{0.62\linewidth}
\centering
\small
\renewcommand\arraystretch{1.1}
\begin{tabular}{c|l|cc}
\hline
\textbf{Method} & \textbf{Model(s)} & \textbf{JGA} & \textbf{JTA} \\
\hline
DS-DST & BERT$_{base}$ & 51.70 & - \\
Seq2Seq-DU & BERT$_{base}$ & 54.40 & - \\
LUNA & BERT$_{base}$ & 56.13 & - \\
\hline
SPACE-3 & UniLM & 57.50 & - \\
\hline
SPLAT & Longformer$_{base}$ & 56.60 & - \\
& Longformer$_{large}$ & 57.40 & - \\
\hline
Diable & T5v1.1$_{base}$ & 56.48 & - \\
D3ST & T5$_{base}$ & 56.10 & - \\
& T5$_{large}$ & 54.20 & - \\
& T5$_{xxl}$ & 58.70 & - \\
TOATOD & T5$_{small}$ & 61.92 & -\\
& T5$_{base}$ & 63.79 & -\\
\hline
\multirow{4}{*}{\makecell{Single\\Expert$_{(ours)}$}} 
& BERT$_{base}$ \; + GNN \quad\quad\quad\quad\;\;\;+ Llama 3.1 8B & 62.66 & 97.66 \\
& BERT$_{base}$ \; + GNN \quad\quad\quad\quad\;\;\;+ Claude 3.7 S. & 64.34 & 97.74 \\
& \quad\quad\quad\quad\quad\;\; T5$_{small}$ \quad\quad\quad\;\; + Llama 3.1 8B & 63.39 & 97.44 \\
& \quad\quad\quad\quad\quad\;\; T5$_{small}$ \quad\quad\quad\;\; + Claude 3.7 S. & 64.66 & 97.50 \\
\hline
\multirow{2}{*}{\makecell{GEM$_{(ours)}$}} 
& (BERT$_{base}$ \;+ GNN) / T5$_{small}$\;\;\;+ Llama 3.1 8B & \textbf{63.57} & 97.56 \\
& (BERT$_{base}$ \;+ GNN) / T5$_{small}$\;\;\;+ Claude 3.7 S. & \textbf{65.19} & 97.65 \\
\hline
\end{tabular}
\caption{This table presents comparative results of our proposed GEM (Graph-Enhanced Mixture) framework against existing SOTA dialogue state tracking models. GEM achieves superior JGA of $65.19\%$ by efficiently combining GNN and T5 architectures with ReAct agent integration.}
\label{tab:moe_performance}
\end{minipage}%
\hfill
\begin{minipage}{0.36\linewidth}
\centering
\small
\begin{tabular}{lcccc}
\hline
& \multicolumn{2}{c}{\textbf{Val}} & \multicolumn{2}{c}{\textbf{Test}} \\
\cline{2-5}
\textbf{Model} & \textbf{Intent} & \textbf{Slot} & \textbf{Intent} & \textbf{Slot} \\
\hline
BERT & 87.46 & 72.67 & 86.84 & \textbf{71.07} \\
BGE-s & 87.36 & 70.36 & 87.00 & 69.71 \\
BGE-b & 86.38 & 69.50 & 85.47 & 68.05 \\
DistilUSE & 82.68 & 51.71 & 81.99 & 52.17 \\
\hline
\end{tabular}
\caption{Embedding model comparison.}
\label{tab:embedding_models}

\vspace{3em}

\small
\begin{tabular}{lcccc}
\hline
& \multicolumn{2}{c}{\textbf{Val}} & \multicolumn{2}{c}{\textbf{Test}} \\
\cline{2-5}
\textbf{$\alpha$:$\beta$:$\gamma$} & \textbf{Intent} & \textbf{Slot} & \textbf{Intent} & \textbf{Slot} \\
\hline
1:1:1 & 86.34 & 73.58 & 86.10 & 72.25 \\
1:0.5:1 & 87.09 & 74.80 & 86.63 & 72.99 \\
1:0.5:2 & 86.85 & 75.86 & 86.53 & \textbf{74.54} \\
1:0.5:3 & 86.64 & 76.09 & 86.39 & 74.39 \\
1:0.5:5 & 86.49 & 75.20 & 86.19 & 73.86 \\
\hline
\end{tabular}
\caption{Loss weighting schemes.}
\label{tab:loss_weights}
\end{minipage}
\vspace{-1em}
\end{table*}

\textbf{Domain-Aware Routing Mechanism:}
As was explained, we introduce a routing architecture that leverages domain-specific characteristics to direct dialogue turns to the most appropriate expert model. Table~\ref{tab:moe_domain} presents the domain-specific routing performance across each domain. The results show an interesting pattern where the router predominantly selects GNN for attraction $91.31\%$ and train $93.48\%$ domains, while favoring T5 for hotel $89.99\%$, restaurant $85.82\%$, and taxi $87.27\%$ domains. This routing strategy reflects the inherent structural differences between domains. Attraction and train domains typically involve more complex entity relationships and key components to extract, which benefits from GNN's graph-structured reasoning capabilities. These domains often contain interconnected information about locations, schedules, and facilities that can be effectively modeled as graphs. Conversely, hotel, restaurant, and taxi domains tend to involve more straightforward slot-value pairs where T5's sequential processing excels.
\\
\textbf{Dialog State Tracking using MoE:}
Table~\ref{tab:moe_performance} presents our proposed methods against baselines on the MultiWOZ dataset. We compare our approach with several SOTA baselines including DS-DST, Seq2Seq-DU, and LUNA (all using BERT$_{base}$ with 110M parameters), SPACE-3 (UniLM with $\sim$340M parameters), SPLAT (using both Longformer$_{base}$ and Longformer$_{large}$), Diable (T5v1.1$_{base}$), D3ST (T5 variants), and TOATOD (T5$_{small}$ and T5$_{base}$), which achieved strong results on the MultiWOZ leaderboard. Among the traditional baselines, TOATOD$_{base}$ achieved the highest performance at $63.79\%$ JGA with 220M parameters, while D3ST$_{xxl}$ achieved $58.70\%$ JGA at the substantial cost of 11B parameters. In contrast, our single models (T5$_{small}$ and GNN) achieve considerably higher performance with dramatically lower parameter counts. The GNN-based models with Llama 3.1 8B and Claude 3.7 Sonnet as value generators achieve $62.66\%$ and $64.34\%$ JGA respectively, while maintaining JTA scores above $97.6\%$ with only 271M parameters. Our T5$_{small}$ architecture achieves $63.39\%$ JGA with Llama 3.1 8B and $64.66\%$ JGA with Claude 3.7 Sonnet, both outperforming the previous SOTA by significant margins while using less than $1\%$ of the parameters compared to D3ST$_{xxl}$ (70M vs 11B). These results demonstrate the effectiveness of our method and the agentic approach proposed in Section~\ref{sec:react}. Our GEM framework combining GNN and T5 experts with ReAct agent-based value generation further improves performance across both LLM families, achieving $63.57\%$ JGA with Llama 3.1 8B and $65.19\%$ JGA with Claude 3.7 Sonnet, demonstrating effective routing between complementary expert architectures through our domain-weighted voting strategy described in Algorithm~\ref{algo:domain-routing}, which dynamically routes dialogue turns based on conversational characteristics.

\subsection{Analysis and Discussion}
Our experimental results demonstrate that specialized architectural components with dynamic routing mechanisms significantly outperform pure LLM approaches for dialogue state tracking, while maintaining computational efficiency through selective expert activation.
\\
\textbf{Performance Analysis:}
As shown in Table~\ref{tab:llm_performance}, the MoE approach significantly outperforms end-to-end LLM methods, achieving 65.19\% JGA compared to 32.50\% JGA for zero-shot Claude (a 32.69 percentage point improvement). Our domain-specific routing analysis reveals that GNN models excel in domains with complex relational structures (attractions and train domains), while T5 performs better in domains with straightforward slot-value extraction patterns (hotels, restaurants, and taxis). The choice of LLM for value generation (Table~\ref{tab:moe_performance}) shows modest but consistent impact, with Claude 3.7 Sonnet achieving $1.62$ percentage points higher JGA than Llama 3.1 8B in the MoE configuration (65.19\% vs 63.57\%) , though the ReAct framework provides the most substantial gains for Llama 3.1 8B.
\\
\textbf{Computational Efficiency:}
Our MoE framework achieves SOTA performance while maintaining computational efficiency through selective expert activation, using either the lightweight T5-Small (70M parameters) or GNN architecture (271M parameters), dramatically fewer than previous approaches like D3ST-T5XXL (11B parameters). As presented in Table~\ref{tab:moe_performance}, while Claude 3.7 Sonnet achieves marginally better performance (65.19\% vs. 63.57\% JGA in MoE configuration), Llama 3.1 8B offers significant advantages in deployment latency and computational cost with only a $1.62$ percentage point JGA trade-off, enabling practical deployment scenarios where cost-effective, low-latency inference is prioritized.

\section{Ablation Studies}
\label{appendix:ablation}
To systematically evaluate and optimize the GNN component of our architecture, we conducted comprehensive ablation studies examining how different GNN configurations contribute to slot detection performance. We focus on four critical aspects: (1) embedding model selection, (2) GNN architecture configuration, (3) dialogue context window size, and (4) loss weighting strategies. All experiments use the MultiWOZ 2.2 dataset with consistent evaluation protocols to ensure optimal GNN performance before integration into our broader MoE framework.

\textbf{Impact of Embedding Models:} We investigated different pre-trained language models for utterance encoding, as shown in Table~\ref{tab:embedding_models}. BERT-base consistently outperforms other models, particularly for slot accuracy. While BGE-small~\cite{xiao2024c} shows comparable intent detection, it underperforms on slot accuracy. DistilUSE-multilingual \cite{reimers-2019-sentence-bert} performs significantly worse on both metrics. These results suggest BERT-base's representations better capture the semantic information required for DST.


\textbf{GNN Architecture Configuration:} We investigated how different GNN configurations affect performance using our BERT-base embedding approach with full dialogue context. We denote configurations as (Layers, Heads, Dimension). In configurations with residual connections (denoted as 'w. residual'), each GAT layer's input is added to its output, facilitating gradient flow and preserving information from earlier layers in deeper architectures. Table~\ref{tab:gat_config} presents results for three configurations of increasing complexity. The results suggest that intent detection benefits from the more compact Baseline configuration, while slot prediction accuracy increase as model size increasing. For our following experiments, we adopted the Large configuration that balances parameter efficiency and model performance, prioritizing slot prediction accuracy which has greater impact on the overall JGA.

\begin{table}[t]
\small
\centering
\begin{tabular}{lcccc}
\hline
& \multicolumn{2}{c}{\textbf{Validation}} & \multicolumn{2}{c}{\textbf{Test}} \\
\cline{2-5}
\textbf{Configuration} & \textbf{Intent} & \textbf{Slot} & \textbf{Intent} & \textbf{Slot} \\
\hline
Small (1L, 4H, 128d) & 86.76 & 70.98 & 86.67 & 69.38 \\
Baseline (2L, 8H, 256d) & 87.46 & 72.67 & 86.84 & 71.07 \\
Large (3L, 12H, 512d) & 87.67 & 73.11 & 86.63 & 71.55 \\
XLarge (4L, 16H, 768d) & 87.51 & 72.29 & 86.54 & 70.67 \\
\hline
Small (1L, 4H, 128d) r. & 87.50 & 71.71 & 86.75 & 69.70 \\
Baseline (2L, 8H, 256d) r. & 87.52 & 73.38 & \textbf{87.03} & 71.28 \\
Large (3L, 12H, 512d) r. & 86.66 & 73.88 & 86.75 & 72.15 \\
XLarge (4L, 16H, 768d) r. & 87.33 & 75.29 & 86.63 & \textbf{73.33} \\
\hline
\end{tabular}
\caption{Performance comparison of different GNN configurations. 'r.' denotes with residual connections.}
\label{tab:gat_config}
\vspace{-1em}
\end{table}

\begin{table}[t]
\small
\centering
\begin{tabular}{lcccc}
\hline
& \multicolumn{2}{c}{\textbf{Validation}} & \multicolumn{2}{c}{\textbf{Test}} \\
\cline{2-5}
\textbf{Context Window} & \textbf{Intent} & \textbf{Slot} & \textbf{Intent} & \textbf{Slot} \\
\hline
0 (current turn only) & 76.93 & 50.38 & 75.61 & 50.08 \\
1 turn & 84.50 & 64.70 & 83.79 & 63.85 \\
2 turns & 85.77 & 70.27 & 84.93 & 69.18 \\
3 turns & 86.13 & 71.79 & 85.09 & 70.43 \\
5 turns & 86.15 & 73.73 & 85.53 & 71.76 \\
10 turns & 86.34 & 73.58 & 86.10 & \textbf{72.25} \\
Full history & 86.66 & 73.88 & 86.75 & 72.15 \\
\hline
\end{tabular}
\caption{Performance comparison of different context window sizes.}
\label{tab:context_window}
\vspace{-1em}
\end{table}

\textbf{Context Window Size:}
Using BERT-base encodings and GNN-Large configuration, we systematically varied the dialogue context window size to quantify its impact on performance. The empirical results shown in Table~\ref{tab:context_window} demonstrate the critical importance of dialogue context for accurate state tracking. The most substantial improvements occur within the first two turns. This dramatic improvement confirms that DST fundamentally requires cross-turn reasoning.

\textbf{Loss Weighting Strategies:} We investigated different weighting schemes for the multi-task learning objective, using a normalized weighted sum formulation:

\begin{equation}
L_{total} = \frac{\alpha \times L_{intent} + \beta \times L_{domain} + \gamma \times L_{slot}}{\alpha + \beta + \gamma},
\end{equation}


\noindent where $\alpha$, $\beta$, and $\gamma$ represent the weights for intent detection, domain classification, and slot detection tasks respectively. Table~\ref{tab:loss_weights} shows performance across different configurations where performance peaks with configuration (1:0.5:2) which achieves $74.54\%$ slot accuracy on the test set. This confirms our hypothesis that slot prediction represents a more challenging task requiring additional optimization emphasis.

\begin{table}[t]
\centering
\small
\begin{tabular}{lcc}
\hline
\textbf{Model} & \textbf{Params} & \textbf{Latency (ms)}\\
\hline
BERT$_{base}$ + GNN & 271M & 5 \\
T5$_{small}$ & 70M & 21 \\
\hline
GEM & 341M & 12 \\
\hline
\end{tabular}
\caption{Comparison of latency per turn for slot extraction models.}
\label{tab:slot_extraction_latency}
\vspace{-1em}
\end{table}

\begin{table}[t]
\centering
\small
\begin{tabular}{lc}
\hline
\textbf{ReAct Agent LLM} & \textbf{Latency (ms)} \\
\hline
Llama 3.1 8B & 2050 \\
Claude 3.7 Sonnet & 3950 \\
\hline
\end{tabular}
\caption{Comparison of latency per turn for LLM-based value extraction. Llama 3.1 8B measurements on 4x NVIDIA L40S, Claude 3.7 Sonnet via Amazon Bedrock API}
\label{tab:llm_latency}
\vspace{-1em}
\end{table}

\section{Latency Analysis}
\label{appendix:latency}

We analyze the computational efficiency of our proposed models by measuring inference latencies across different components of our system, as shown in Tables~\ref{tab:slot_extraction_latency} and~\ref{tab:llm_latency}. For slot extraction efficiency, our individual expert models demonstrate highly optimized performance with BERT$_{base}$ + GNN achieving 5ms per turn with 271M parameters, while T5$_{small}$ provides 21ms per turn with only 70M parameters, representing different points on the efficiency-capacity trade-off curve for slot extraction tasks. Our GEM method successfully combines both expert architectures through an optimized gating mechanism, achieving a balanced 12ms per turn latency with 341M total parameters, demonstrating that the mixture approach provides a practical middle ground between the speed of BERT-based processing and the generative capabilities of T5 while maintaining deployment feasibility for real-time dialogue systems. Regarding LLM value generation trade-offs, we observe substantial differences in inference performance where Llama 3.1 8B deployed on 4x NVIDIA L40S GPUs achieves 2050ms per turn compared to Claude 3.7 Sonnet's 3950ms per turn, representing a 48\% latency reduction that makes Llama 3.1 8B particularly attractive for latency-sensitive applications where the modest 1.62 percentage point decrease in Joint Goal Accuracy is an acceptable trade-off for nearly doubling the inference speed.

\section{Conclusion and Future Work}
In this work, we present GEM, a novel Mixture-of-Experts framework for Dialogue State Tracking that addresses the fundamental limitations of end-to-end LLM approaches in structured information extraction. By dynamically routing between a Graph Neural Network that captures dialogue structure and a T5 encoder-decoder for sequence modeling, followed by slot-value extraction using an LLM-based ReAct approach, our method achieves 65.19\% Joint Goal Accuracy on MultiWOZ 2.2, outperforming SOTA methods like TOATOD (63.79\%). Our analysis reveals that specialized architectures excel at capturing dialogue dynamics that general-purpose LLMs struggle with, particularly in multi-domain conversations requiring precise slot-value tracking.

Future work will extend the framework to cross-dataset evaluation, explore modeling the MoE router using additional dimensions such as dialogue text and chains of domains / intents / slots from previous turns, and investigate knowledge distillation into smaller unified models for resource-constrained deployment. These directions advance more efficient and deployable dialogue understanding systems.







\bibliography{aaai2026}

\newpage
\appendix

\clearpage

\end{document}